\documentclass[journal]{IEEEtran}
\usepackage{cite}
\usepackage{graphicx}

\ifCLASSINFOpdf
\else
\fi
\usepackage{amsmath}
\usepackage{algorithmic}
\usepackage{algorithm}
\usepackage{url}

\usepackage{multirow}
\usepackage{hhline}
\usepackage{array}
\usepackage{amsmath}
\usepackage{amssymb}
\usepackage{bm}
\usepackage{color}
\usepackage{tabularx}
\newcolumntype{Y}{>{\centering\arraybackslash}X}
\usepackage[colorlinks = true,
            linkcolor = blue,
            urlcolor  = blue,
            anchorcolor = blue]{hyperref}
\definecolor{COLOR}{rgb}{1.0, 0.25, 0.25}

\begin{document}
%
\title{CleftNet: Augmented Deep Learning for Synaptic
Cleft Detection from Brain Electron Microscopy}
%
%
%

\author{Yi~Liu,
        and~Shuiwang~Ji,~\IEEEmembership{Senior Member,~IEEE}
\thanks{Yi Liu and Shuiwang Ji are with the Department
of Computer Science \& Engineering, Texas A\&M University, College Station, TX 77843, USA
(email: yiliu@tamu.edu, sji@tamu.edu).}
\thanks{This work was supported by National Institutes of Health grant 1R21NS102828.}}
\maketitle

\begin{abstract}
Detecting synaptic clefts is a crucial step to investigate the biological function of synapses.
The volume electron microscopy (EM)
allows the identification of synaptic clefts by photoing EM images with high resolution and fine details.
Machine learning approaches have been employed to automatically predict synaptic clefts from EM images.
In this work, we propose a novel and augmented deep learning model, known as CleftNet, for improving 
synaptic cleft detection from brain EM images.
We first propose two novel network components, known as the feature augmentor and the label augmentor,
for augmenting features and labels to improve cleft representations.
The feature augmentor can fuse global information from inputs and learn common morphological patterns in clefts,
leading to augmented cleft features.
In addition, it can generate outputs with varying dimensions, making it flexible to be integrated
in any deep network.
The proposed label augmentor augments the label of each voxel from a value to a vector,
which contains both the segmentation label and boundary label.
This allows the network to learn important shape information and to produce more informative cleft representations.
Based on the proposed feature augmentor and label augmentor, We build the CleftNet as a U-Net like network.
The effectiveness of our methods is evaluated on both online and offline tasks.
Our CleftNet currently ranks \#1 on the online task of the CREMI open challenge.
In addition, both quantitative and qualitative
results in the offline tasks show that our method outperforms the baseline
approaches significantly.
\end{abstract}


%
\IEEEpeerreviewmaketitle

\section{Introduction}
Synapses are fundamental biological structures that transmit
signals across neurons.
A synapse is composed of several pre-synaptic neurons, a pro-synaptic neuron, 
and a synaptic cleft between these two types of neurons.~\cite{becker2013learning, buhmann2019automatic}.
Detecting synaptic clefts is a key step to 
reconstruct synapses and
investigate synaptic functions.
Currently, the volume electron microscopy (EM) is recognized as 
the most reliable technique for reconstruction of neural circuits~\cite{lee2019convolutional,Li:TMI17,Fakhry:TMI,Zeng:Bioinfo17,Fakhry:bioinfo16,wei2020mitoem}.
It can provide 3D EM images with high resolution and sufficient details 
for synaptic cleft detection and synaptic structure analysis~\cite{dorkenwald2019binary,dorkenwald2017automated}.

The manual annotation of synaptic clefts
is time-consuming and requires heavy labor from domain experts~\cite{zheng2018complete}.
This raises the need of constructing
computational models to automatically detect synaptic clefts from EM images.
Machine learning approaches~\cite{becker2013learning,kreshuk2011automated,dorkenwald2019binary,heinrich2018synaptic} have been employed to learn relationships
between EM images and the annotated clefts.
Then for any newly obtained EM image, the synaptic clefts it contains could be directly predicted by the trained models.
Earlier studies~\cite{becker2013learning,kreshuk2011automated} employ traditional machine learning methods
for synaptic cleft detection.
These methods require
carefully-designed features based on
the prior knowledge of synapses and EM images.

With the rapid development of deep learning, recent methods~\cite{dorkenwald2019binary,heinrich2018synaptic} 
automatically learn features from EM images using deep neural networks.
Thus, hand-crafted features are avoided and models can be learned end-to-end.
Importantly, synaptic cleft detection is formulated as an image dense prediction problem,
where each voxel in the input volume is predicted as a cleft voxel or not.
There exist several deep architectures for the dense prediction of biological images~\cite{milletari2016v,long2015fully,ronneberger2015u,christiansen2018silico,christ2016automatic,wang2018interactive}, among which networks based-on the
U-Net architecture~\cite{ronneberger2015u,cciccek20163d,Liu:TMI2020,wang2020global}
are shown to achieve the best performance.
Existing studies~\cite{dorkenwald2019binary,heinrich2018synaptic}
apply vanilla U-Nets to cleft detection and obtain the state-of-the-art results.
However, they do not consider the unique properties of clefts, \emph{e.g.}, 
morphological patterns and geometrical shapes of clefts.

In this work, we propose the CleftNet, a novel deep learning model 
for improving synaptic cleft detection from brain EM images.
Our CleftNet contains two novel components, known as the feature augmentor (FA) and the label augmentor (LA),
to augment features and labels by considering and learning important properties of clefts.
The feature augmentor
is inspired by the attention mechanisms~\cite{oktay2018attention,vaswani2017attention}.
The feature representation
of each output position in the FA is dependent on features of all positions in the input.
Hence, global information is captured effectively to improve cleft features.
Importantly, a trainable query tensor is used to learn common morphological patterns of all synaptic clefts.
The query is fully trainable and updated by all input EM images. Thus, it can capture common patterns and learn topological structures in clefts, leading to augmented feature representations.
Note that our proposed FA is flexible in terms of generating outputs of different sizes.
Thus, it can simulate any commonly-used operations, like pooling, convolution, or deconvolution, and can be easily integrated
into any deep neural network.

In addition to augmenting features using the FA, we propose the 
label augmentor to augment labels for voxels.
Existing methods mainly use original segmentation labels, which is demonstrated to
be biased towards learning image textures while ignoring shape information~\cite{takikawa2019gated,geirhos2018imagenet}.
However, such information is important to circuit reconstruction based on
biological images~\cite{heinrich2018synaptic}. To this end,
We proposed the LA to augment the label of each voxel from a scalar to a vector, which contains the segmentation label
and the boundary label for the voxel.
By doing this, the network is encouraged to learn both the texture and shape information from inputs,
leading to more informative representations.
Finally, we build the CleftNet, a U-Net like network integrating the proposed feature augmentors
and label augmentors. We replace all the pooling layers, deconvolution layers, and the bottom block
with the appropriately sized FAs to generate augmented features.
A novel loss function based on the LA is used in the CleftNet
for learning more powerful cleft representations. 

We conduct comprehensive experiments to evaluate our methods.
First, we apply our methods to the MICCAI Challenge on Circuit Reconstruction from Electron Microscopy Images (CREMI). Our CleftNet now ranks \#1 on the synaptic cleft detection task of the CREMI open challenge.
Next, we compare our methods with several baseline methods on both the validation set and the test data.
Quantitative and qualitative results show that our methods achieve significant improvements over the baselines.
In addition, we conduct ablation studies to demonstrate the effectiveness of the important network components including the FA, LA and the learnable query.
Overall, our contributions are summarized as below:
\begin{itemize}
    \item We propose the feature augmentor that improves cleft features
    through fusing global information from inputs and learning shared patterns in clefts.
    It can generate outputs with varying dimensions and can be adapted to any deep network with high flexibility.
    \item We design the label augmentor that augments the label for each voxel from a scalar to a vector.
    The LA enables the network to prioritize the learning of shape information, which is important for
    identifying synaptic clefts.
    \item We develop the CleftNet, an augmented deep learning method considering and learning specific morphological patterns and shape information in clefts. CleftNet generates improved cleft representations by incorporating the proposed feature augmentor
    and label augmentor.
    \item CleftNet is the new state-of-the-art model for synaptic cleft detection.
    It not only ranks \#1 on the online task of the CREMI open challenge, but also outperforms
    other baseline methods significantly on offline tasks.
\end{itemize}

\section{Related Work}
\subsection{Synaptic Cleft Detection}
Machine learning methods have been explored for synaptic cleft detection.
Earlier studies~\cite{becker2013learning,kreshuk2011automated} employ traditional machine learning methods
with carefully-designed features. The work in~\cite{becker2013learning} designs features for synapses
by considering similar texture cues shared by clefts. The constructed features are then taken by an AdaBoost-based classifier to
identify synapses and clefts.
The work in~\cite{kreshuk2011automated} applies a random-forest classifier to 
hand-crafted features that constructed based on the geometrical information in EM volumes.
Recently, deep learning methods~\cite{dorkenwald2019binary,heinrich2018synaptic} are used to
automatically learn features from EM images.
Synaptic cleft detection is formulated as
an image dense prediction problem in these methods.
The work in~\cite{dorkenwald2019binary} uses the 3D residual U-Net~\cite{zhang2018road} to 
detect cleft voxels from a newly obtained large EM dataset. 
The misalignment issue of adjacent patches is tackled by designing a training strategy that is robust to misalignments.
The work in~\cite{heinrich2018synaptic} employs the
vanilla 3D U-Net with an auxiliary task to detect clefts and uses the CREMI~\cite{Anu:2013} dataset.
Both approaches apply well-studied deep architectures and obtain state-of-the-art detection performance 
on the used datasets. In this work, we propose a novel deep model, known as CleftNet, to improve synaptic cleft detection by augmenting cleft features and labels of clefts.

\begin{figure*}[t]
    \centering
    \includegraphics[width=\textwidth]{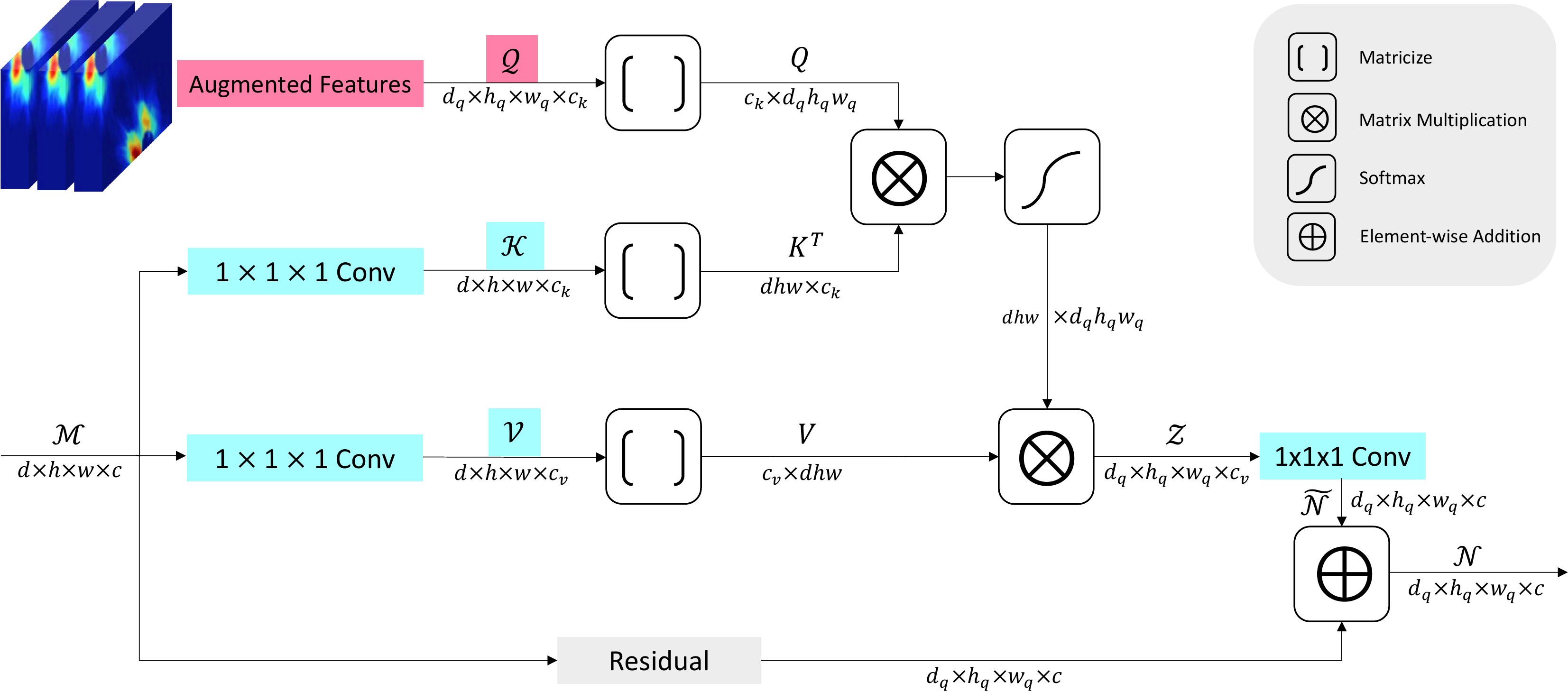}
    \caption{An illustration of the proposed feature augmentor (FA)
    as detailed in Section~\ref{sec:fa}.
    For each operation,
    the generated tensor/matrix and the dimensions are marked aside the corresponding arrow.
    The operations for converting tensors back to matrices are not included for simplification.
    The query tensor $\mathcal{Q}$ contains free parameters and is trained during the whole learning process
    to capture shared patterns of cleft features. The depth, height, and the width of the output tensor $\mathcal{N}$
    are determined by those of $\mathcal{Q}$. These dimensions can be flexibly adapted per design requirements. Hence,
    the proposed FA can replace
    any commonly-used operation like pooling or deconvolution, and
    can be
    integrated to any deep architecture with high flexibility.
    }\label{fig:ca}
\end{figure*}

\subsection{Attention on Images}
We summarize the attention-based methods on images in this section.
Generally, there exist two categories of attention methods for images in literature;
those are, gated-attention (GateAttn) methods and self-attention (SelfAttn) methods.
We start by defining annotations for clear illustration.
Given an input image tensor $\mathcal{M}$, we convert it
to a matrix $M\in\mathbb{R}^{c\times s}$ following appropriate modes as introduced in~\cite{kolda2009tensor}, where $s$ denotes the spatial dimensions, and $c$ denotes the channel dimension.
For instance, $s=dhw$ if the input is a 3D image, where $d$, $h$ and $w$
denote the depth, height, and the width, respectively.

\subsubsection{Gated-attention Methods}
The GateAttn aims at selecting important features from inputs using a trainable vector.
The selection could be conducted along the spatial direction, namely spatial-wise attention (SWA), 
as used in the spatial attention module in CBAM~\cite{chen2020improved} and the
attention U-Net~\cite{oktay2018attention}.
In the SWA, the matrix $M\in\mathbb{R}^{c\times s}$ could be treated as a set of vectors as
$\{\mathbf{m}_i \in\mathbb{R}^{c}\}^{s}_{i=1}$, where each $\mathbf{m}_i$ is
a vector representation for the pixel or voxel $i$, and $i=1,...,s$.
If the selection is performed along the channel direction, it is known as the channel-wise attention (CWA) and is used in
the channel attention module in CBAM~\cite{chen2020improved} and the squeeze and excitation networks~\cite{hu2018squeeze}.
We use the matrix $M^T\in\mathbb{R}^{s\times c}$ in CWA and it could be treated as a set of vectors as $\{\tilde{\mathbf{m}}_j \in\mathbb{R}^{s}\}^{c}_{j=1}$, where each $\tilde{\mathbf{m}}_j$ is 
a vector representation for the channel $j$, and $j=1,...,c$.

Let $\mathbf{q}_s\in\mathbb{R}^{c}$ and $\mathbf{q}_c\in\mathbb{R}^{s}$ denote the learnable vectors
for the SWA and the CWA.
The corresponding importance vectors $\mathbf{a}_s$ and $\mathbf{a}_c$ are computed as
\begin{equation}\label{eq:sattn}
\begin{aligned}
&\mathbf{a}_s = \mbox{Softmax}(M^T\mathbf{q}_s)\in\mathbb{R}^{s},\\
&\mathbf{a}_c = \mbox{Softmax}(M\mathbf{q}_c)\in\mathbb{R}^{c},\\
\end{aligned}
\end{equation}
where $\mbox{Softmax}()$ denotes the element-wise softmax operation.
Essentially, each $a^i_s$ in $\mathbf{a}_s$ is the inner product of $\mathbf{m}_i$ and $\mathbf{q}_s$, which serves as the 
important score for $\mathbf{m}_i$.
Similarly, the important score $a^j_c$ for $\tilde{\mathbf{m}}_j$ is computed as the inner product of $\tilde{\mathbf{m}}_j$ and $\mathbf{q}_c$.
Finally, in SWA, each $\mathbf{m}_i$ is scaled by its important score $a^i_s$,
resulting in the output matrix $N_s$ as a set of vectors as $\{a^i_s\mathbf{m}_i \in\mathbb{R}^{c}\}^{s}_{i=1}$.
Similarly, in CWA, the generated matrix $N_c$ could be treated as $\{a^j_c\tilde{\mathbf{m}}_j \in\mathbb{R}^{s}\}^{c}_{i=1}$.
Essentially, SWA imposes an importance score for each pixel/voxel vector representation, and
CWA applies an importance score for each channel vector representation.
The obtained $N_s$ or $N_c$ can be converted back to the output tensor that has the same dimensions as the input $\mathcal{M}$. 
Importantly, the GateAttn is used to highlight important pixels/voxels or channels,
but fails in capturing global information to the output.
In addition, using a trainable vector constraints the network's capability to learn complicated patterns,
such as the common morphological patterns shared by all clefts.

\subsubsection{Self-attention Methods}
The SelfAttn is firstly introduced in the work~\cite{vaswani2017attention} and then
applied to images~\cite{zhang2019self} and videos~\cite{wang2018non}.
It first performs linear transformation on the $\mathcal{M}$ three times and generates three tensors;
those are, the query $\mathcal{Q}$,
the key $\mathcal{K}$, and the value $\mathcal{V}$. These three tensors are converted to three matrix $Q$, $K$, and $V$ along the appropriate modes~\cite{kolda2009tensor}. Each matrix has the dimensions of $c\times s$ same as $M$.
SelfAttn is then performed as
\begin{equation}\label{eq:selfattn}
\begin{aligned}
N = V \cdot\mbox{Softmax}(K^TQ)\in\mathbb{R}^{c\times s},
\end{aligned}
\end{equation}
where $\mbox{Softmax}()$ denotes the column-wise softmax operation.
$N$ is then converted back to the tensor $\mathcal{N}$ that has the same dimensions as $\mathcal{M}$.
By performing SelfAttn, 
Each pixel/voxel feature in the output $\mathcal{N}$ is dependent on features of all the pixels/voxels in the input $\mathcal{M}$.
To this end, 
SelfAttn is used to capture long-range dependencies and aggregate global information from inputs.
However, all the three tensors $\mathcal{Q}$, $\mathcal{K}$, and $\mathcal{V}$ are generated from inputs, making them input-dependent,
which may not able to capture the common patterns shared by all clefts.
In addition, SelfAttn preserves the dimensions from inputs, 
which leads to its nature of low flexibility.
For instance, we can only replace a convolution with a stride of 1 with SelfAttn.
It can not simulate other operations that change the input dimensions,
such as pooling or deconvolution.

\section{The Proposed Methods}

\subsection{Feature Augmentation} \label{sec:fa}
The self-attention mechanism~\cite{vaswani2017attention,wang2018non} has
achieved great success in various domains, including natural
language processing~\cite{yang2016hierarchical,LiuTextICDM19} and
computer vision~\cite{Liu:TMI2020,wang2020global,zwangUnetAAAI20}.
Based upon the self-attention mechanism, we propose a novel method, known
as the feature augmentor (FA), to improve cleft feature
representations. Our proposed FA not only aggregates global
information from the whole feature space of inputs, but also
captures shared patterns of all synaptic clefts in the whole
dataset, thereby leading to augmented and improved feature
representations.

An illustration of our proposed FA is provided in
Figure~\ref{fig:ca}. Formally, let
$\mathcal{M}\in\mathbb{R}^{d\times h\times w\times c}$ denote the
input tensor to the FA, where $d$ is the depth, $h$ is the
height, $w$ is the width, and $c$ is the number of feature maps. We
first generate the key tensor $\mathcal{K}$ and the value tensor
$\mathcal{V}$ from the input tensor $\mathcal{M}$ by performing two
separate 3D convolutions, each of which has a kernel size of $1
\times 1 \times 1$ and a stride of $1$. The used convolutions
retain the spatial sizes of the input tensor but generate varying
numbers of feature maps. Assuming the first convolution above produces
$c_k$ feature maps, thus, we have $\mathcal{K}\in\mathbb{R}^{d\times
h\times w\times c_k}$. Similarly, we obtain
$\mathcal{V}\in\mathbb{R}^{d\times h\times w\times c_v}$ assuming
the second convolution generates $c_v$ feature maps. Note that
$\mathcal{K}$ and $\mathcal{V}$ are both generated from
$\mathcal{M}$, hence, they are input-dependent.

Next, we consider the query tensor $\mathcal{Q}$ in our proposed FA.
Existing studies~\cite{Liu:TMI2020,wang2020global,zwangUnetAAAI20}
use the same strategies to generate
$\mathcal{Q}$ as $\mathcal{K}$ and $\mathcal{V}$, making it to be input-dependent.
In this work, we instead formulate $\mathcal{Q}\in\mathbb{R}^{d_q\times h_q\times w_q\times c_q}$ as a learnable tensor containing
free parameters, where $d_q$, $h_q$, $w_q$, and $c_q$
denote the depth, height, width, and the number of feature maps, respectively.
The parameters in $\mathcal{Q}$ are randomly initialized and trained along with other parameters
during the whole learning process.
Note that $d_q$, $h_q$, $w_q$ can be any positive integers, but
$c_q$ needs to be equal to
 $c_k$ as required by the attention mechanism.

The main purpose of using a learnable query is to capture shared patterns of all synaptic clefts.
Even though different synaptic clefts exhibit distinct shapes,
they share common morphological patterns.
Such patterns are important for learning structural topology
and improving cleft feature representations.
Existing studies~\cite{Liu:TMI2020,wang2020global,zwangUnetAAAI20} obtain the query tensor by imposing a linear transformation on the input tensor $\mathcal{M}$. However,
they suffer from two limitations when learning cleft features.
First, the input-dependent query fails to capture the common patterns shared by all clefts.
In addition,
the shared patterns are complicated, and it is difficult to explicitly capture them by performing a linear transformation on the input.
To this end,
we use a trainable query $\mathcal{Q}$ to automatically learn such complicated and shared patterns.
As $\mathcal{Q}$ is shared and updated by all input volumes in the dataset, it is expected to grasp common patterns and help to augment
cleft features.

After obtaining the tensors $\mathcal{K}$, $\mathcal{V}$, and $\mathcal{Q}$,
we perform attention by first converting them to three matrices. Formally, we have
\begin{equation}\label{eq:unfold}
\begin{aligned}
&K = \mbox{Matricize}(\mathcal{K})\in\mathbb{R}^{c_k\times dhw},\\
&V = \mbox{Matricize}(\mathcal{V})\in\mathbb{R}^{c_v\times dhw },\\
&Q = \mbox{Matricize}(\mathcal{Q})\in\mathbb{R}^{c_k\times d_qh_qw_q},\\
\end{aligned}
\end{equation}
where $\mbox{Matricize}()$ denotes the mode-4 matricization of an input tensor~\cite{kolda2009tensor}.
The attention is then performed as
\begin{equation}\label{eq:att}
\begin{aligned}
Z = V \cdot\mbox{Softmax}(K^TQ)\in\mathbb{R}^{c_v\times d_qh_qw_q},
\end{aligned}
\end{equation}
where $\mbox{Softmax}()$ denotes the column-wise softmax operation.
The $Z$ is then converted back to the tensor $\mathcal{Z}\in\mathbb{R}^{d_q \times h_q \times w_q\times c_v}$
along mode-4~\cite{kolda2009tensor}.
Another $1 \times 1 \times 1$ 3D convolution is performed on
$\mathcal{Z}$ to produce $\tilde{\mathcal{N}}\in\mathbb{R}^{d_q \times h_q \times w_q\times c}$ to
match the number of feature maps of the input $\mathcal{M}$.
Then the
final output $\mathcal{N}$ of the proposed FA is computed as
\begin{equation}\label{eq:final}
\begin{aligned}
\mathcal{N} = \tilde{\mathcal{N}} + \mbox{Residual}(\mathcal{M})\in\mathbb{R}^{d_q \times h_q \times w_q\times c},
\end{aligned}
\end{equation}
where $\mbox{Residual}()$ denotes the residual operations that directly map the input to the output with the purpose of
reusing features and accelerating training~\cite{he2016deep}.
It is obvious that the depth, height, and width of $\mathcal{N}$ are determined by $\mathcal{Q}$.
Theoretically, $d_q$, $h_q$, and $w_q$ can be any positive integers.
In practice, we consider three scenarios to be
corresponding to other commonly-used operations.
For instance, we set $d_q = 1/2 d, h_q = 1/2 h, w_q = 1/2 w$ to correspond with a $2 \times 2 \times 2$ pooling or a convolution
with a stride of $2$; we set $d_q = 2 d, h_q = 2 h, w_q = 2 w$ to match up with a deconvolution with a stride of 2;
we can also preserve the spatial dimensions from the input to simulate a convolution with a stride of 1.
For the residual operations, accordingly,
we simply use a $2 \times 2 \times 2$ max pooling,
a trilinear interpolation with a factor of 2,
and an identical mapping for these three scenarios.

Generally, the matrix $Q$ in Eq.~\ref{eq:att} can be treated as a set of query vectors $\{\mathbf{q}_i \in\mathbb{R}^{c_q}\}^{d_qh_qw_q}_{i=1}$, and it is similar to $K$, $V$ and the produced $Z$.
$K^TQ$ enables each query $\mathbf{q}_i$ to attend each key vector $\mathbf{k}_j$, generating an attention map as a matrix $(a_{ji})$, where $i=1,...,d_qh_qw_q$, and $j=1,...,dhw$.
This matrix is
further normalized by performing the column-wise softmax operation.
Next, by multiplying $V$ at the left side,
each output vector $\mathbf{z}_h$ in $Z$ is obtained as a weighted sum of all the value vectors in $V$,
with each $\mathbf{v}_g$ scaled by the weight $a_{gh}$ in the attention map matrix.
By doing this, the feature vector of each voxel in the tensor $\mathcal{Z}$ is dependent on feature vectors of all
the voxels in the input.
Hence, global information from the whole feature space of the input is captured effectively.
In addition, we use a learnable query tensor $\mathcal{Q}$ to automatically learn shared pattern features
of all synaptic clefts. The query $\mathcal{Q}$ is randomly initialized and trained during the whole learning process. As $\mathcal{Q}$ is shared and updated by all input volumes, it is expected to grasp common patterns and shape structural relationships in clefts,
thereby leading to
augmented feature representations.

\begin{figure}[t]
    \centering
    \includegraphics[width=0.45\textwidth]{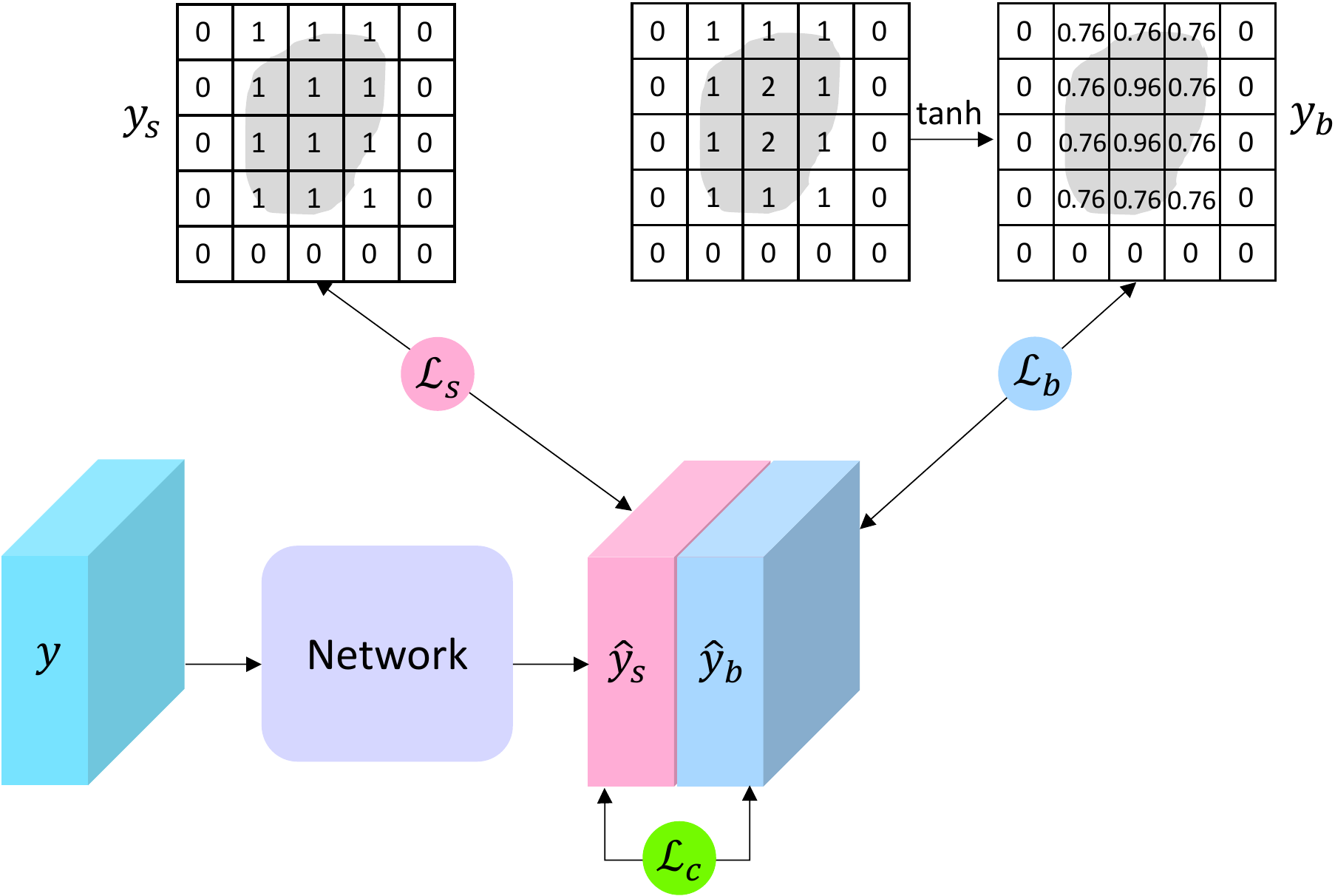}
    \caption{An illustration of the proposed label augmentor (LA) as detailed in Section~\ref{sec:la}.
    For each voxel $i$ of the input volume $y$, LA augments its label to a vector $[y^i_{s}, y^i_{b}]^T$, where
    $y^i_{s}$ is the segmentation label and $y^i_{b}$ is the boundary label.
    The gray areas indicate the same cleft.
    The boundary labels $y_b$ are computed using tanh distance map (TDM) as shown in the figure.
    The same network with weight sharing is employed to learn predictions on the two sets of labels.
    The segmentation loss $\mathcal{L}_{s}$ for the volume $y$ is achieved from the segmentation predictions $\hat{y}_s$
    and segmentation labels $y_s$. The boundary loss $\mathcal{L}_{b}$ is calculated from the boundary predictions $\hat{y}_b$
    and boundary labels $y_b$. The coherence loss $\mathcal{L}_{c}$ is computed based on the divergences between predictions
    $\hat{y}_s$ and $\hat{y}_b$.
    }\label{fig:la}
\end{figure}

\subsection{Label Augmentation} \label{sec:la}

Deep neural networks (DNNs)
have achieved state-of-the-art performance on the tasks of dense prediction for 3D images~\cite{cciccek20163d,Chen:KDD18,wang2020global}
Existing methods mainly perform voxel-wise classification
and the label information only contains the category for each voxel.
In this work, we propose the label augmentor (LA)
to augment labels and explicitly consider shape information of clefts.

Generally, an image contains color, shape, and texture information.
These information is combined together to DNNs for pixel/voxel-wise classification~\cite{takikawa2019gated,karimi2019reducing}.
The work in~\cite{geirhos2018imagenet} demonstrates that
DNNs have strong biases towards learning textures but pay little attention to shapes.
In biological-image analysis, however, manual annotations from domain experts are usually
conducted with the prior knowledge of boundary and shape identification.
For instance, segmenting clefts from tissue volumes would entail following cleft boundaries and
consequently narrowing the region of interest.
Hence, it is natural to leverage shape information to enhance the capability of the networks.
To this end, we propose to prioritize the learning of boundary information by augmenting labels.
Specifically, for each voxel, we augment its label from a scalar to a vector, which consists of
both the segmentation category and the corresponding boundary information.
The proposed label augmentor enables the multi-task learning that captures ample information from input EM images.

There exist several methods to compute and incorporate boundary information~\cite{takikawa2019gated,xue2020shape,dangi2019distance,ni2019elastic}.
In this work, we follow the thread and use
tanh distance map (TDM) to compute the tanh of the Euclidean distance of each voxel from
the nearest boundary voxel~\cite{borgefors1986distance}.
Formally, let $\Lambda$ denote the set of cleft voxels for one synaptic cleft, and let
$\partial \Lambda$ represent the set of boundary voxels for this synaptic cleft. Then the tanh distance map, $y_{b}$, is defined as
\begin{equation}\label{eq:sdm}
y^i_{b} =\left\{
\begin{array}{lr}
 \mbox{tanh}(\min \limits_{j \in \partial \square} \mbox{d}(i, j)) & i \in \Lambda \\
 0 & i \notin \Lambda\\
\end{array}
\right.,
\end{equation}
where $i$ is the index of
any voxel in the input volume, $\mbox{d}(\ , \ )$ denotes
the Euclidean distance for any two voxels,
and $\mbox{tanh}()$ is the element-wise tanh operation.
Apparently, $y^i_{b}$ is in the range of $(0,1)$ based on the definition.
Essentially, TDM reveals how each voxel contributes
to the shape of the corresponding cleft.
Let $y^i_{s}$ denote the original segmentation category of the voxel $i$.
To this end, we augment the label for each voxel $i$ to a vector
\boldmath$\ell$\unboldmath$^i = [y^i_{s}, y^i_{b}]^T$.

We use the same network to produce the outputs of two streams corresponding to two labels, and the parameters are
shared across both streams.
By doing this, the network is forced to learn rich texture and shape information,
thereby leading to more powerful representations.
Let $y^i_{s}=1$ indicate the voxel $i$ is a cleft voxel.
We use the weighted binary cross-entropy loss for the segmentation steam as
\begin{equation}\label{eq:loss_sef}
\begin{aligned}
\mathcal{L}_{s} = - \beta_{s} \sum_{i\in y^+_{s}} \mbox{log}P(y^i_{s}=1)-(1-\beta_{s})\sum_{i\in y^-_{s}} \mbox{log}P(y^i_{s}=0),
\end{aligned}
\end{equation}
where $\beta_{s}$ is the ratio of non-cleft voxels to all voxels in the volume,
$P(y^i_{s})$ is the probability of the predicted class for the voxel $i$,
and $y^+_{s}$, $y^-_{s}$ are the sets of indexes of voxels whose segmentation labels are clefts, non-clefts, respectively.
As the data is imbalanced and the non-clefts are overwhelmingly dominant in the volume,
$\beta_{s}$ is close to one while $1-\beta_{s}$ is close to zero.
Intuitively, we use larger penalties for cleft voxels in the loss function to tackle
the imbalance issue.

Essentially, the segmentation stream is a voxel-wise classification problem,
while the boundary stream is a voxel-wise regression problem.
We use the weighted L2 loss for the boundary stream as
\begin{equation}\label{eq:loss_bound}
\begin{aligned}
\mathcal{L}_{b} = \beta_{b} \sum_{i\in y^+_{b}} (y^i_{b}-\hat{y}^i_{b})^2+(1-\beta_{b})\sum_{i\in y^-_{b}} (y^i_{b}-\hat{y}^i_{b})^2,
\end{aligned}
\end{equation}
where $\beta_{b}$ is the ratio of voxels
whose boundary labels are larger than zero to all voxels in the volume,
$\hat{y}^i_{b}$ is the predicted boundary distance of the voxel $i$,
and $y^+_{b}$, $y^-_{b}$ are the sets of indexes of voxels whose
boundary labels are positive, and zeros, respectively.

We augment the label of each voxel $i$ to a vector \boldmath$\ell$\unboldmath$^i = [y^i_{s}, y^i_{b}]^T$,
and the boundary label $y^i_b$ is computed based on the provided segmentation label $y^i_s$.
Importantly, there exists correspondence between these two scalars.
Specifically, the segmentation label $y^i_s=1$ is in line with
the boundary label $y^i_b>0$, and  $y^i_s=0$ corresponds to
$y^i_b=0$.
By augmenting labels and
employing the same network with weight sharing, the texture and shape information are jointly learned together in the training phase.
However, there may still exist mismatch in the predicted segmentation map and shape map.
Here, we propose a coherence loss as a regularizer as
\begin{equation}\label{eq:loss_cohen}
\begin{aligned}
\mathcal{L}_{c} = -\sum_{i\in \hat{y}^+_{b}} (1-\hat{y}^i_{s})\mbox{log}\hat{y}^i_{b}- \sum_{i\in \hat{y}^-_{b}} (1-\hat{y}^i_{b})\mbox{log}P(y^i_{s}=1), \\
\end{aligned}
\end{equation}
where $\hat{y}^+_{b}$, and $\hat{y}^-_{b}$denote the sets of voxels whose predicted boundary values are positive, and non-positive,
respectively.
Intuitively, we impose a penalty to voxel $i$ if the two predicted values
$\hat{y}^i_{s} $ and $ \hat{y}^i_{b}$
are not consistent with each other.
Specifically, there are two terms in Eq.~\ref{eq:loss_cohen}.
The former penalizes the voxel $i$ whose predicted boundary value is positive while the predicted segmentation value is zero.
We use the function $-\mbox{log}$ to impose heavier penalties for the voxels close to the boundary.
For the latter, the voxel is penalized if the predicted boundary value is zero but the predicted segmentation value is one.
Note that $\hat{y}^i_{b}$ ranges from $0$ to $1$ in the former. Correspondingly, we use $P(y^i_{s}=1)$
to produce a probability that also ranges from $0$ to $1$ in the latter.

The final loss is computed as the weighted sum of the above three loss functions as
\begin{equation}\label{eq:loss_final}
\begin{aligned}
\mathcal{L} = \mathcal{L}_{s} + \alpha_1\mathcal{L}_{b} + \alpha_2\mathcal{L}_{c},
\end{aligned}
\end{equation}
where the weights $\alpha_1$ and $\alpha_2$ are hyper-parameters.
By doing this, the network is forced to learn high-level patterns for both the segmentation label and boundary label.
As a result, shape information is explicitly considered in LA to benefit synaptic cleft detection.

\subsection{CleftNet}
Deep architectures have been intensively studied for dense prediction of
biological images~\cite{long2015fully,ronneberger2015u,Liu:TMI2020},
among which networks based on the U-Net architecture are shown to achieve
superior performance~\cite{ronneberger2015u,cciccek20163d,Liu:TMI2020}.
Essentially, the U-Net consists of an encoder and a decoder, which are connected by
the bottom block and several skip connections.
In this work, we propose the CleftNet, a U-Net like network with the proposed feature augmentors
and label augmentors, for improving synaptic cleft detection from EM images.
Specifically, we replace all the downsampling layers in the encoder,
all the upsampling layers in the decoder, and the bottom block of the vanilla U-Net with the corresponding
FAs as introduced in Section~\ref{sec:fa}.
For example, we replace a downsampling layer with a FA where the depth, height, and width of the learnable
query $\mathcal{Q}$ are all set to be half of those of the input volume.
In addition, the residual operation used in this FA is a
$2 \times 2 \times 2$ max pooling.
Essentially, The proposed FA augments cleft features by aggregating
global information from the whole input volume and capturing commonly shared patterns of
all clefts in the whole dataset.
By incorporating the proposed LA as described in Section~\ref{sec:la}, the CleftNet outputs a vector with two elements for each voxel corresponding to its label vector
containing the segmentation label and the boundary label.
The final loss function of the CleftNet is the weighted sum of the three loss functions as mentioned in Eq.~\ref{eq:loss_final}.
By doing this, the network is forced to learn both the texture and the shape information for
synaptic clefts, thereby leading to more powerful cleft representations.

\section{Experimental Studies}

\subsection{Dataset} \label{sec:data}
We use the dataset from the MICCAI Challenge on Circuit Reconstruction from Electron Microscopy Images (CREMI)~\cite{Anu:2013}.
The CREMI dataset was obtained from the adult \emph{Drosophila} using serial section transmission electron microscopy (ssTEM) with the resolution $40 \mbox{nm} \times 4 \mbox{nm}\times 4 \mbox{nm}$.
The training data is consist of three volumes A, B, and C, each of which has
the spatial sizes of $125 \times 1250 \times 1250$.
The original segmentation label for the training data indicates whether a voxel is a background voxel or a cleft voxel.
There are also three volumes A+, B+, and C+ with the same dimensions as the training volumes in the test data, but
the label is not publicly provided.
The CREMI challenge requires the submitting of predicted results to conduct unbiased
comparisons among different methods.
Generally, the data is sparse regarding the ratio of cleft voxels to total voxels.


\subsection{Network Settings}
Our proposed CleftNet follows the general settings in the 3D U-Net architecture~\cite{cciccek20163d}
and ResUnet~\cite{zhang2018road}
with integrating our proposed FAs and LAs.
Both the encoder and decoder contain 4 blocks,
and the numbers of output channels of the 4 block are 32, 64, 96, 128, respectively.
The bottom block outputs 160 channels.
We replace the second convolution block in each of the 9 blocks in the original 3D U-Net architecture~~\cite{cciccek20163d} with a residual block.
By doing this,
each of the 9 blocks contains a convolution block, a residual block, and a corresponding FA, sequentially.
Each of the 4 used FAs in the encoder serves as a downsampling layer such that
the spatial sizes are halved along the depth, height, and the width directions.
In the decoder, the FA can be used for performing upsampling and recovering
the spatial sizes of inputs.
Convolution layers and residual blocks in the encoder and decoder are used for extracting features.
Each convolution block is consist of a convolution layer with kernel $3 \times 3 \times 3$ and stride 1, 
followed by a batch normalization (BN) layer and ELu with $\alpha$ equal to 1~\cite{clevert2015fast}.
For each residual block,
we employ the original version of ResNets in~\cite{he2016identity} that
addition is performed after the second BN layer and before the second Elu.
The output block integrates our proposed LA and generates two channels for both the textual and shape learning.
The weights in Eq.~\ref{eq:loss_final} are $\alpha_1= 0.5$, $\alpha_2= 0.2$, respectively.

\subsection{Training and Inference}
Each of the three volumes A, B, C in the training data contains 125 slices.
We use the first 100 slices of all the A, B, C as the training set.
The last 25 slices of A, B, C are used as the validation set.
In this way, the training set contains three volumes with sizes of
$100 \times 1250 \times 1250$, while
the validation set contains three volumes with sizes of
$25 \times 1250 \times 1250$.

We compute the augmented label vector based on the original segmentation label for each voxel of the input volumes.
The input sizes of the network are set to be $8 \times 256 \times 256$.
During training, a 3D patch with sizes $8 \times 256 \times 256$ is randomly cropped from the any of the three
volumes as input to the network to train the weights. During inference, we first perform prediction for each patch and then stack all the predicted patches together to generate the final prediction map for each volume.
As the data is sparse, we further tackle the imbalance issue
by counting the number of cleft voxels in an input patch.
Specifically, we reject a patch with a high probability of 95\% if
it contains less than 200 cleft voxels.
Our data augmentation techniques include
rotation with probability 0.5, flip with probability 0.5, and grayscale with probability 0.2.
We set the batch size to be 16 using four NVIDIA GeForce RTX 2080 Ti GPUs.
To optimize the model,
we employ the Adam optimizer~\cite{kingma2014adam} with a learning
rate of 0.001.
All hyper-parameters are tuned on the validation set.
We run $10e5$ iterations in total, and inference on
the validation set is performed per 100 iterations.

After obtaining the optimized model,
we conduct inference on the test data, which contains three volumes
A+, B+, and C+ with sizes of $125 \times 1250 \times 1250$.
We use the same input patch sizes $8 \times 256 \times 256$ and generate predicted patches
then stack them together as the final predictions.
As the ground truth of the test data is not publicly provided,
we submitted the predictions to CREMI open challenge to compute
the model performance.

\begin{table}
    \caption{Results on the task of synaptic cleft detection
    of the CREMI open challenge.
    We summarize the results of the best models for the top 5 groups.
    Our group is DIVE, and our best model is CleftNet.
    A downward arrow $\downarrow$ is used to indicate
    that a lower metric value
    corresponds to better performance.
    These results are provided at the CREMI website~\cite{Anu:2013}.
    }\label{table:leader}
    \begin{tabularx}{\linewidth}{l Y c Y}\hline
        \textbf{Rank}  & Group & Model      & \textbf{CREMI-score}($\downarrow$)    \\ \hline\hline
        1       & DIVE        & CleftNet          & 57.73          \\ \hline
        4      & CleftKing        & CleABC1          & 58.02          \\ \hline
        6        & lbdl        & lb\_thicc            & 59.04          \\ \hline
        9        & 3DEM         & BesA225          & 59.34        \\ \hline
        15               & SDG         & Unet2           & 63.92          \\ \hline
        \hline
    \end{tabularx}
\end{table}

\subsection{Metrics}
As data is imbalanced regarding the ratio of cleft voxels to total voxels in three volumes,
we use F1-score, AUC, and CREMI-score for evaluation.
Specifically, AUC represents area under the ROC curve, where the ROC curve is created
by plotting true positive rate (TPR) against the false positive rate (FPR).
Formally,
$\mbox{TPR} = \frac{\mbox{TP}}{\mbox{TP}+\mbox{FN}}$, and
$\mbox{FPR} = \frac{\mbox{FP}}{\mbox{FP}+\mbox{TN}}$,
where TP, FN, FP, and TN represent true positive, false negative, false positive, and true negative, respectively.
In addition, for the F1-score and CREMI-score, we have
\begin{equation}\label{eq:metrics}
\begin{aligned}
&\mbox{F1-score} = \frac{2}{\mbox{recall}^{-1} + \mbox{precision}^{-1}},\\
&\mbox{CREMI-score} = \frac{\mbox{ADGT} + \mbox{ADF}}{2},
\end{aligned}
\end{equation}
where $\mbox{recall}=\frac{\mbox{TP}}{\mbox{TP}+\mbox{FN}}$, and $\mbox{precision}=\frac{\mbox{TP}}{\mbox{TP}+\mbox{FP}}$.
ADGT represents the average distance of any predicted cleft voxel to the closest ground truth cleft voxel, which
is related to FPs in predictions.
ADF denotes the average distance of any ground truth cleft voxel to the closest predicted cleft voxel, which is related to FNs in predictions.
Essentially, F1-score is the Harmonic mean of recall and precision,
and CREMI-score aims at measuring the overall distance
between a predicted cleft and a true cleft.
Note that CREMI-score is used as the primary ranking metric on the leaderboard of synaptic cleft detection of the CREMI open challenge.

\begin{table}
    \caption{Comparison among different models in terms of CREMI-score, F1-Score, and AUC
    on the validation set.
    An upward arrow $\uparrow$ is used to indicate
    that a higher metric value
    corresponds to better performance.
    A downward arrow $\downarrow$ is used to indicate
    that a lower metric value
    corresponds to better performance.
    The best performance is in bold.
    }\label{table:overall_val}
    \begin{tabularx}{\linewidth}{l Y c Y}\hline
        \textbf{Model}        & \textbf{CREMI-score}($\downarrow$)    & \textbf{AUC}($\uparrow$)  & \textbf{F1-score}($\uparrow$)  \\ \hline\hline
        SegNet       & 72.32         & 0.883           & 0.842           \\ \hline
        3D U-Net       & 64.27         & 0.875           & 0.837          \\ \hline
        V-Net        & 65.73        & 0.896            & 0.858          \\ \hline
        ResUnet        & 59.64          & 0.909           & 0.874         \\ \hline
        AttnUnet               & 57.34         & 0.914            & 0.878          \\ \hline
        \textbf{CleftNet} & \textbf{50.50} & \textbf{0.938}   & \textbf{0.895}  \\ \hline
        \hline
    \end{tabularx}
\end{table}

\begin{table}
    \caption{Comparison among different models in terms of CREMI-score and F1-Score
    on the test data. The best performance is in bold.
    }\label{table:overall_test}
    \begin{tabularx}{\linewidth}{l Y c}\hline
        \textbf{Model}        & \textbf{CREMI-score}($\downarrow$) & \textbf{F1-score}($\uparrow$) \\ \hline\hline
        SegNet       &82.46  & 0.794                                \\ \hline
        V-Net        & 76.12  & 0.744                          \\ \hline
        3D U-Net     & 75.02  & 0.800                              \\ \hline
        ResUnet      & 66.50   & 0.813                            \\ \hline
        AttnUnet      & 65.19         & 0.812                            \\ \hline
        \textbf{CleftNet} & \textbf{57.73}  & \textbf{0.831}  \\ \hline
        \hline
    \end{tabularx}
    \vspace{-8pt}
\end{table}

\subsection{CREMI Open Challenge Task}
The MICCAI Challenge on Circuit Reconstruction from Electron Microscopy Images (CREMI)~\cite{Anu:2013}
started from the year of 2016 and has become one of the most well-known open challenges on EM data.
There are totally three tasks in the CREMI challenge including neuron segmentation, synaptic cleft detection, and
synaptic partner identification.
We focus on the task of synaptic cleft detection and the used data is described in Section~\ref{sec:data}.
As the label for the test data is not publicly provided,
we submitted the predicted results of our CleftNet
to the task of synaptic cleft detection of the CREMI challenge, where CREMI-score is used as the primary ranking metric.
There are at least 15 groups participating in this task.
We summarize the results of the best models for the top 5 groups in Table~\ref{table:leader}.
Our best model CleftNet ranks the first with the CREMI-score of 57.73.
Note we do not use any model ensemble technique to enhance the performance.

\subsection{Comparison with Baselines}
We compare our methods with several state-of-the-art architectures
including SegNet~\cite{badrinarayanan2017segnet}, V-Net~\cite{milletari2016v}, 3D U-Net~\cite{cciccek20163d},
ResUnet~\cite{zhang2018road}, and AttnUnet~\cite{oktay2018attention}.
To ensure fair comparisons, we use similar architectures for all methods,
including number of blocks for the encoders and decoders,
numbers of output features of each block, etc.
In addition, we integrate their official code for
specific components (like residual blocks and attention units)
to the architectures.

For the validation set, we directly report the computed values for all metrics
from the optimized models.
The comparison results are provided in Table~\ref{table:overall_val}.
We can observe from the table that our proposed CleftNet
consistently outperforms other baseline methods on all the three metrics.
Specifically, compared with the best baseline method AttnUnet, CleftNet refines the CREMI-score by a large margin of
6.84, which is a considerable value of 13.5\% of our current CREMI-score 50.50.
CleftNet reduces the CREMI-score by an average margin of 13.76 when
considering all the five baseline methods, which is an incredible percentage of 27.2\% of our current score.
In addition, CleftNet outperforms the baselines on the other two metrics
by average margins of 4.25\% in terms of AUC and 3.72\% in terms of F1-score.
All of these results demonstrate the effectiveness of CleftNet with the proposed feature augmentors and label augmentors.

For the test data,
we submitted the predicted volumes from all the optimized models to the CREMI open challenge for evaluation.
The results in terms of CREMI-score and F1-score are provided at the CREMI website~\cite{Anu:2013} and
summarized in Table~\ref{table:overall_test}.
We draw similar conclusions that CleftNet performs the best on the test data.
Specifically, CleftNet outperforms the other five methods by an average margin of 15.33 in terms of CREMI-score,
which is 26.6\% of the current score 57.73.
In addition,
CleftNet achieves an average improvement of 3.12\% in terms of F1-score.
The CREMI open challenge does not provide results for AUC.

\begin{table}
    \caption{Comparison among ResUnet, AttnUnet, CleftNet without feature augmentors (FAs),
    CleftNet without label augmentor (LAs), and CleftNet
    in terms of CREMI-score, F1-Score, and AUC
    on the validation set.
    }\label{table:aba_val}
    \begin{tabularx}{\linewidth}{l Y c Y}\hline
        \textbf{Model}        & \textbf{CREMI-score}($\downarrow$)    & \textbf{AUC}($\uparrow$)  & \textbf{F1-score}($\uparrow$)  \\ \hline\hline
        ResUnet        & 59.64          & 0.909           & 0.874           \\ \hline
        AttnUnet               & 57.34         & 0.914            & 0.878          \\ \hline
        CleftNet w/o FA & 53.37 & 0.922   & 0.883  \\ \hline
        CleftNet w/o LA & 52.65 & 0.930   & 0.890  \\ \hline
        CleftNet & 50.50 & 0.938   & 0.895  \\ \hline
        \hline
    \end{tabularx}
\end{table}

\begin{table}
    \caption{Comparison among ResUnet, AttnUnet, CleftNet without feature augmentor (FA),
    CleftNet without label augmentor (LA), and CleftNet
    in terms of CREMI-score and F1-Score
    on the test data.
    }\label{table:aba_test}
    \begin{tabularx}{\linewidth}{l Y c Y}\hline
        \textbf{Model}        & \textbf{CREMI-score}($\downarrow$)     & \textbf{F1-score}($\uparrow$)  \\ \hline\hline
        ResUnet        & 66.50                & 0.813          \\ \hline
        AttnUnet               & 65.19                 & 0.812          \\ \hline
        CleftNet w/o FA & 59.11  & 0.820  \\ \hline
        CleftNet w/o LA & 59.03    & 0.823  \\ \hline
        CleftNet & 57.73    & 0.831  \\ \hline
        \hline
    \end{tabularx}
\end{table}

\subsection{Ablation Studies}
We conduct ablation study to
investigate the effectiveness of our proposed components feature augmentor (FA) and label augmentor (LA).
We remove the FAs from CleftNet which we denote as CleftNet w/o FA, and
we remove the LAs from CleftNet which we denote as CleftNet w/o LA.
If we remove both the FAs and LAs then CleftNet would simply become ResUnet.
We also include the results of AttnUnet to compare the attention units used in AttnUnet and our proposed FA.
Experimental results on the validation set is provided in Table~\ref{table:aba_val}.
We can observe from the table that CleftNet outperforms ResUnet by large margins of
9.14, 2.7\%, and 2.1\% in terms of CREMI-score, AUC, and F1-score, respectively.
As we use ResUnet as the backbone architecture, the better performance achieved by
CleftNet compared with ResUnet indicates the effectiveness of our proposed FA ans LA.
In addition, the contribution of FAs can be shown from two comparisons; these are,
the better results of CleftNet w/o LA over ResUnet,
and the better results of CleftNet over CleftNet w/o FA.
Similarly, both the better performances of
CleftNet w/o FA over ResUnet and CleftNet over
CleftNet w/o LA reveal the effectiveness of LAs.

Notably, the only difference of CleftNet w/o LA from AttnUnet
is the use of the FAs rather that the attention units proposed in AttnUnet.
Table~\ref{table:aba_val} shows that compared with AttnUnet,
CleftNet w/o LA induces improvements of 6.99, 2.1\%, and 1.6\%
in terms of CREMI-score, AUC, and F1-score, respectively.
This reveals the superior capability of our proposed FA than the attention unit.
Essentially,
the used attention unit in AttnUnet employs a query vector to highlight more important
voxels in the input. The inner product between the query vector and each voxel vector produces a
scalar as the importance score of the corresponding voxel.
However, it fails to aggregate global context information
from the whole input to the output, imposing natural constraints to the network's capability.
In addition, a single vector may not sufficient to store and learn the commonly shared
patterns of all clefts in the dataset.
Our FA uses a learnable query tensor with high dimensions and generates
response of each voxel by considering all the voxels in the input.
By doing this, FA not only extracts global information from the whole input volume,
but also captures the complicated shared patterns of all clefts in the dataset,
leading to augmented and improved cleft features.

We summarize the results on the test data in Table~\ref{table:aba_test}.
For CREMI score, CleftNet w/o LA reduces the score by 7.47 compared with ResUnet,
and CleftNet reduces the score by 1.38 compared with CleftNet w/o FA.
Both of these demonstrate the contribution of the proposed FA.
Similarly, the effectiveness of the LA can be demonstrated by either the improvement of 7.39
induced by CleftNet w/o FA over ResUnet, or the the improvement of 1.30 induced by
CleftNet over CleftNet w/o LA.
Essentially, we draw consistent conclusions
that both the FA and LA make considerable contributions for learning better cleft representations.

\begin{table}
    \caption{Comparison between CleftNet w Self-attention (SelfAtt) and
    CleftNet with feature augmentor (FA)
    in terms of CREMI-score, F1-Score, and AUC
    on the validation set.
    }\label{table:query}
    \begin{tabularx}{\linewidth}{l Y c Y}\hline
        \textbf{Model}        & \textbf{CREMI-score}($\downarrow$)    & \textbf{AUC}($\uparrow$)  & \textbf{F1-score}($\uparrow$)  \\ \hline\hline
        CleftNet w SelfAttn & 51.98 & 0.932   & 0.894  \\ \hline
        CleftNet w FA & 50.50 & 0.938   & 0.895  \\ \hline
        \hline
    \end{tabularx}
\end{table}

\begin{figure*}[t]
    \centering
    \includegraphics[width= \textwidth]{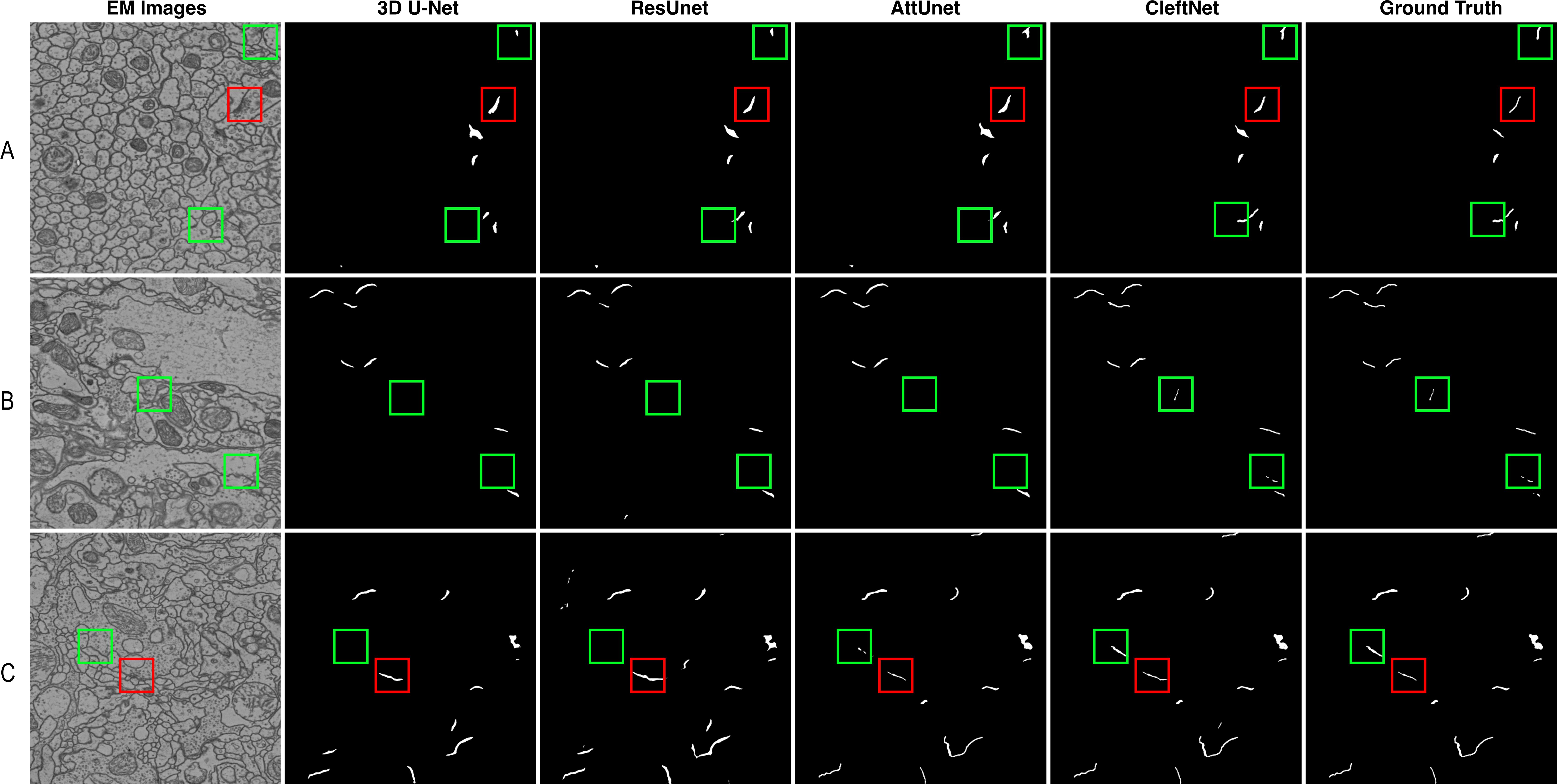}
    \caption{Qualitative results for 3D U-Net, ResUnet, AttnUnet, and CleftNet
    on the validation set. Each row shows results for a slice from the corresponding volume
    A, B or C.
    The red boxes mark the regions where false positives are easily produced in predictions.
    The green boxes mark the regions where false negatives could be easily generated in predictions.
    }\label{fig:vis}
\end{figure*}

\subsection{Investigating Different Queries}
We investigate the effectiveness of the learnable query in FA compared the
input-dependent query in self-attention~\cite{vaswani2017attention}.
The studies~\cite{Liu:TMI2020,wang2020global} use similar strategies to integrate
attention-based methods in U-Net,
but the query tensor is achieved by
performing a linear transformation on the input. By doing this, the query is similar as the key and value
that to be input-dependent. For clear illustration, we denote our method as CleftNet w FA, and
the methods in~\cite{Liu:TMI2020,wang2020global} as CleftNet w SelfAttn.
We conduct experiments to compare these two methods and the results on the validation set are reported
in Table~\ref{table:query}.
We can observe from the table that our method induces slight improvements based on
CleftNet w SelfAttn.
Specifically, CleftNet w FA outperforms CleftNet w SelfAtt by margins of
1.48, 0.6\%, and 0.1\% in terms of CREMI-score, AUC, and F1-score, respectively.
This indicates that compared with using self-attention, integrating our proposed FA in U-Net generates
better cleft representations.
The performance improvements are achieved by purely using the learnable query rather than
the input-dependent query.
Note that our FA can achieve one similar goal as self-attention that capturing global context information.
The output response of each voxel fuses information from all the voxels in the whole input volume.
However, commonly shared patterns for clefts are complicated, and it is difficult to
obtain these patterns by performing a linear transformation on the input.
The self-attention used in~\cite{Liu:TMI2020,wang2020global} designs the query to be input-dependent,
failing to store and learn complicated inherent patterns shared by all the clefts in the dataset.
Instead, in our FA, we design a trainable query tensor to
automatically learn such complex patterns during the whole training process and produce better cleft representations.

\subsection{Qualitative Results}
We compare the four best methods among six qualitatively by showing
visualization results in Figure~\ref{fig:vis}.
Specifically, we compare 3D U-Net, ResUnet, AttnUnet, and CleftNet on the validation set
for clear comparisons
as the validation set has labels.
We show the predicted results for three slices, and each of them is from
the volume A, B, and  C, respectively.
The figure evidently shows that CleftNet generates best cleft predictions among
the four methods.
Generally, some non-cleft cracks in tissues appear to be similar as clefts,
and it is possible that these voxels are predicted as cleft voxels,
thereby resulting in false positives.
In addition, some clefts have small sizes that the contained voxels can be easily predicted as
background voxels, which inducing false negatives.
Our CleftNet produces the least false positives or false negatives,
generating predictions that are closest to the ground truth.
This again demonstrates the effectivenesses of the proposed feature augmentor and label augmentor in CleftNet.

\section{Conclusion}
In this work, 
we propose a novel deep network, known as CleftNet, for synaptic cleft detection from brain EM images.
Our CleftNet is a U-Net like network with the purpose of learning specific properties of synaptic clefts,
such as common morphological patterns and shape information of clefts.
We achieve this goal by augmenting the network with two network components, 
the feature augmentor and the label augmentor.
The feature augmentor is designed based on the self-attention mechanism but 
uses a learnable query and generates outputs with arbitrary spatial dimensions.
It can not only capture global information from inputs, but also learn complicated morphological
patterns shared by all clefts, leading to aumgmented cleft features.
The feature augmentor can replace any commonly-used operation, such as convolution, pooling, or deconvolution.
Hence, it can be integrated into any deep model with high flexibility.
The original label for each input voxel is the segmentation label purely, and 
we propose the label augmentor to augment it to a vector,
which contains both the segmentation label and boundary label.
This essentially enables the multi-task learning, and the network is capable of learning
both the texture and shape information of clefts, generating
more informative representations.
We integrate these two components and propose the CleftNet,
the new state-of-the-art model for synaptic cleft detection.
CleftNet not only ranks first on the online task of the CREMI challenge,
but also performs much better than the baseline methods on offline tasks.

%
%

\begin{thebibliography}{10}
\providecommand{\url}[1]{#1}
\csname url@samestyle\endcsname
\providecommand{\newblock}{\relax}
\providecommand{\bibinfo}[2]{#2}
\providecommand{\BIBentrySTDinterwordspacing}{\spaceskip=0pt\relax}
\providecommand{\BIBentryALTinterwordstretchfactor}{4}
\providecommand{\BIBentryALTinterwordspacing}{\spaceskip=\fontdimen2\font plus
\BIBentryALTinterwordstretchfactor\fontdimen3\font minus
  \fontdimen4\font\relax}
\providecommand{\BIBforeignlanguage}[2]{{%
\expandafter\ifx\csname l@#1\endcsname\relax
\typeout{** WARNING: IEEEtran.bst: No hyphenation pattern has been}%
\typeout{** loaded for the language `#1'. Using the pattern for}%
\typeout{** the default language instead.}%
\else
\language=\csname l@#1\endcsname
\fi
#2}}
\providecommand{\BIBdecl}{\relax}
\BIBdecl

\bibitem{becker2013learning}
C.~Becker, K.~Ali, G.~Knott, and P.~Fua, ``Learning context cues for synapse
  segmentation,'' \emph{IEEE transactions on medical imaging}, vol.~32, no.~10,
  pp. 1864--1877, 2013.

\bibitem{buhmann2019automatic}
J.~Buhmann, A.~Sheridan, S.~Gerhard, R.~Krause, T.~Nguyen, L.~Heinrich
  \emph{et~al.}, ``Automatic detection of synaptic partners in a whole-brain
  drosophila em dataset,'' \emph{bioRxiv}, 2019.

\bibitem{lee2019convolutional}
K.~Lee, N.~Turner, T.~Macrina, J.~Wu, R.~Lu, and H.~S. Seung, ``Convolutional
  nets for reconstructing neural circuits from brain images acquired by serial
  section electron microscopy,'' \emph{Current opinion in neurobiology},
  vol.~55, pp. 188--198, 2019.

\bibitem{Li:TMI17}
R.~Li, T.~Zeng, H.~Peng, and S.~Ji, ``Deep learning segmentation of optical
  microscopy images improves {3D} neuron reconstruction,'' \emph{IEEE
  Transactions on Medical Imaging}, vol.~36, no.~7, pp. 1533--1541, 2017.

\bibitem{Fakhry:TMI}
A.~Fakhry, T.~Zeng, and S.~Ji, ``Residual deconvolutional networks for brain
  electron microscopy image segmentation,'' \emph{IEEE Transactions on Medical
  Imaging}, vol.~36, no.~2, pp. 447--456, 2017.

\bibitem{Zeng:Bioinfo17}
T.~Zeng, B.~Wu, and S.~Ji, ``{DeepEM3D}: Approaching human-level performance on
  {3D} anisotropic {EM} image segmentation,'' \emph{Bioinformatics}, vol.~33,
  no.~16, pp. 2555--2562, 2017.

\bibitem{Fakhry:bioinfo16}
A.~Fakhry, H.~Peng, and S.~Ji, ``Deep models for brain {EM} image segmentation:
  novel insights and improved performance,'' \emph{Bioinformatics}, vol.~32,
  pp. 2352--2358, 2016.

\bibitem{wei2020mitoem}
D.~Wei, Z.~Lin, D.~Franco-Barranco, N.~Wendt, X.~Liu, W.~Yin \emph{et~al.},
  ``Mitoem dataset: Large-scale 3d mitochondria instance segmentation from em
  images,'' in \emph{International Conference on Medical Image Computing and
  Computer-Assisted Intervention}.\hskip 1em plus 0.5em minus 0.4em\relax
  Springer, 2020, pp. 66--76.

\bibitem{dorkenwald2019binary}
S.~Dorkenwald, N.~L. Turner, T.~Macrina, K.~Lee, R.~Lu, J.~Wu \emph{et~al.},
  ``Binary and analog variation of synapses between cortical pyramidal
  neurons,'' \emph{BioRxiv}, 2019.

\bibitem{dorkenwald2017automated}
S.~Dorkenwald, P.~J. Schubert, M.~F. Killinger, G.~Urban, S.~Mikula, F.~Svara
  \emph{et~al.}, ``Automated synaptic connectivity inference for volume
  electron microscopy,'' \emph{Nature methods}, vol.~14, no.~4, pp. 435--442,
  2017.

\bibitem{zheng2018complete}
Z.~Zheng, J.~S. Lauritzen, E.~Perlman, C.~G. Robinson, M.~Nichols, D.~Milkie
  \emph{et~al.}, ``A complete electron microscopy volume of the brain of adult
  drosophila melanogaster,'' \emph{Cell}, vol. 174, no.~3, pp. 730--743, 2018.

\bibitem{kreshuk2011automated}
A.~Kreshuk, C.~N. Straehle, C.~Sommer, U.~Koethe, M.~Cantoni, G.~Knott
  \emph{et~al.}, ``Automated detection and segmentation of synaptic contacts in
  nearly isotropic serial electron microscopy images,'' \emph{PloS one},
  vol.~6, no.~10, p. e24899, 2011.

\bibitem{heinrich2018synaptic}
L.~Heinrich, J.~Funke, C.~Pape, J.~Nunez-Iglesias, and S.~Saalfeld, ``Synaptic
  cleft segmentation in non-isotropic volume electron microscopy of the
  complete drosophila brain,'' in \emph{International Conference on Medical
  Image Computing and Computer-Assisted Intervention}.\hskip 1em plus 0.5em
  minus 0.4em\relax Springer, 2018, pp. 317--325.

\bibitem{milletari2016v}
F.~Milletari, N.~Navab, and S.-A. Ahmadi, ``V-net: Fully convolutional neural
  networks for volumetric medical image segmentation,'' in \emph{2016 fourth
  international conference on 3D vision (3DV)}.\hskip 1em plus 0.5em minus
  0.4em\relax IEEE, 2016, pp. 565--571.

\bibitem{long2015fully}
J.~Long, E.~Shelhamer, and T.~Darrell, ``Fully convolutional networks for
  semantic segmentation,'' in \emph{Proceedings of the IEEE conference on
  computer vision and pattern recognition}, 2015, pp. 3431--3440.

\bibitem{ronneberger2015u}
O.~Ronneberger, P.~Fischer, and T.~Brox, ``U-net: Convolutional networks for
  biomedical image segmentation,'' in \emph{International Conference on Medical
  image computing and computer-assisted intervention}.\hskip 1em plus 0.5em
  minus 0.4em\relax Springer, 2015, pp. 234--241.

\bibitem{christiansen2018silico}
E.~M. Christiansen, S.~J. Yang, D.~M. Ando, A.~Javaherian, G.~Skibinski,
  S.~Lipnick \emph{et~al.}, ``In silico labeling: Predicting fluorescent labels
  in unlabeled images,'' \emph{Cell}, vol. 173, no.~3, pp. 792--803, 2018.

\bibitem{christ2016automatic}
P.~F. Christ, M.~E.~A. Elshaer, F.~Ettlinger, S.~Tatavarty, M.~Bickel, P.~Bilic
  \emph{et~al.}, ``Automatic liver and lesion segmentation in ct using cascaded
  fully convolutional neural networks and 3d conditional random fields,'' in
  \emph{International Conference on Medical Image Computing and
  Computer-Assisted Intervention}.\hskip 1em plus 0.5em minus 0.4em\relax
  Springer, 2016, pp. 415--423.

\bibitem{wang2018interactive}
G.~Wang, W.~Li, M.~A. Zuluaga, R.~Pratt, P.~A. Patel, M.~Aertsen \emph{et~al.},
  ``Interactive medical image segmentation using deep learning with
  image-specific fine tuning,'' \emph{IEEE transactions on medical imaging},
  vol.~37, no.~7, pp. 1562--1573, 2018.

\bibitem{cciccek20163d}
{\"O}.~{\c{C}}i{\c{c}}ek, A.~Abdulkadir, S.~S. Lienkamp, T.~Brox, and
  O.~Ronneberger, ``3d u-net: learning dense volumetric segmentation from
  sparse annotation,'' in \emph{International conference on medical image
  computing and computer-assisted intervention}.\hskip 1em plus 0.5em minus
  0.4em\relax Springer, 2016, pp. 424--432.

\bibitem{Liu:TMI2020}
Y.~Liu, H.~Yuan, Z.~Wang, and S.~Ji, ``Global pixel transformers for virtual
  staining of microscopy images,'' \emph{IEEE Transactions on Medical Imaging},
  vol.~39, no.~6, pp. 2256--2266, 2020.

\bibitem{wang2020global}
Z.~Wang, Y.~Xie, and S.~Ji, ``Global voxel transformer networks for augmented
  microscopy,'' \emph{Nature Machine Intelligence}, 2020.

\bibitem{oktay2018attention}
O.~Oktay, J.~Schlemper, L.~L. Folgoc, M.~Lee, M.~Heinrich, K.~Misawa
  \emph{et~al.}, ``Attention u-net: Learning where to look for the pancreas,''
  \emph{arXiv preprint arXiv:1804.03999}, 2018.

\bibitem{vaswani2017attention}
A.~Vaswani, N.~Shazeer, N.~Parmar, J.~Uszkoreit, L.~Jones, A.~N. Gomez
  \emph{et~al.}, ``Attention is all you need,'' in \emph{Advances in Neural
  Information Processing Systems}, 2017, pp. 6000--6010.

\bibitem{takikawa2019gated}
T.~Takikawa, D.~Acuna, V.~Jampani, and S.~Fidler, ``Gated-scnn: Gated shape
  cnns for semantic segmentation,'' in \emph{Proceedings of the IEEE
  International Conference on Computer Vision}, 2019, pp. 5229--5238.

\bibitem{geirhos2018imagenet}
R.~Geirhos, P.~Rubisch, C.~Michaelis, M.~Bethge, F.~A. Wichmann, and
  W.~Brendel, ``Imagenet-trained cnns are biased towards texture; increasing
  shape bias improves accuracy and robustness,'' in \emph{International
  Conference on Learning Representations (ICLR)}, 2019.

\bibitem{zhang2018road}
Z.~Zhang, Q.~Liu, and Y.~Wang, ``Road extraction by deep residual u-net,''
  \emph{IEEE Geoscience and Remote Sensing Letters}, vol.~15, no.~5, pp.
  749--753, 2018.

\bibitem{Anu:2013}
D.~B. S. T. E.~P. Jan~Funke, Stephan~Saalfeld, ``Miccai challenge oncircuit
  reconstruction from electron microscopy images,'' \url{http://cremi.org/},
  2016.

\bibitem{kolda2009tensor}
T.~G. Kolda and B.~W. Bader, ``Tensor decompositions and applications,''
  \emph{SIAM review}, vol.~51, no.~3, pp. 455--500, 2009.

\bibitem{chen2020improved}
X.~Chen, H.~Fan, R.~Girshick, and K.~He, ``Improved baselines with momentum
  contrastive learning,'' \emph{arXiv preprint arXiv:2003.04297}, 2020.

\bibitem{hu2018squeeze}
J.~Hu, L.~Shen, and G.~Sun, ``Squeeze-and-excitation networks,'' in
  \emph{Proceedings of the IEEE conference on computer vision and pattern
  recognition}, 2018, pp. 7132--7141.

\bibitem{zhang2019self}
H.~Zhang, I.~Goodfellow, D.~Metaxas, and A.~Odena, ``Self-attention generative
  adversarial networks,'' in \emph{International conference on machine
  learning}.\hskip 1em plus 0.5em minus 0.4em\relax PMLR, 2019, pp. 7354--7363.

\bibitem{wang2018non}
X.~Wang, R.~Girshick, A.~Gupta, and K.~He, ``Non-local neural networks,'' in
  \emph{Proceedings of the IEEE conference on computer vision and pattern
  recognition}, 2018, pp. 7794--7803.

\bibitem{yang2016hierarchical}
Z.~Yang, D.~Yang, C.~Dyer, X.~He, A.~Smola, and E.~Hovy, ``Hierarchical
  attention networks for document classification,'' in \emph{Proceedings of the
  2016 Conference of the North American Chapter of the Association for
  Computational Linguistics: Human Language Technologies}, 2016, pp.
  1480--1489.

\bibitem{LiuTextICDM19}
Y.~Liu, H.~Yuan, and S.~Ji, ``Learning local and global multi-context
  representations for document classification,'' in \emph{Proceedings of the
  19th IEEE International Conference on Data Mining}, 2019, pp. 1234--1239.

\bibitem{zwangUnetAAAI20}
Z.~Wang, N.~Zou, D.~Shen, and S.~Ji, ``Non-local {U}-nets for biomedical image
  segmentation,'' in \emph{Proceedings of the 34th AAAI Conference on
  Artificial Intelligence}, 2020, pp. 6315--6322.

\bibitem{he2016deep}
K.~He, X.~Zhang, S.~Ren, and J.~Sun, ``Deep residual learning for image
  recognition,'' in \emph{Proceedings of the IEEE conference on computer vision
  and pattern recognition}, 2016, pp. 770--778.

\bibitem{Chen:KDD18}
Y.~Chen, H.~Gao, L.~Cai, M.~Shi, D.~Shen, and S.~Ji, ``Voxel deconvolutional
  networks for {3D} brain image labeling,'' in \emph{Proceedings of the 24th
  ACM SIGKDD International Conference on Knowledge Discovery and Data Mining},
  2018, pp. 1226--1234.

\bibitem{karimi2019reducing}
D.~Karimi and S.~E. Salcudean, ``Reducing the hausdorff distance in medical
  image segmentation with convolutional neural networks,'' \emph{IEEE
  Transactions on medical imaging}, vol.~39, no.~2, pp. 499--513, 2019.

\bibitem{xue2020shape}
Y.~Xue, H.~Tang, Z.~Qiao, G.~Gong, Y.~Yin, Z.~Qian \emph{et~al.}, ``Shape-aware
  organ segmentation by predicting signed distance maps,'' in \emph{Proceedings
  of the AAAI Conference on Artificial Intelligence}, vol.~34, no.~07, 2020,
  pp. 12\,565--12\,572.

\bibitem{dangi2019distance}
S.~Dangi, C.~A. Linte, and Z.~Yaniv, ``A distance map regularized cnn for
  cardiac cine mr image segmentation,'' \emph{Medical physics}, vol.~46,
  no.~12, pp. 5637--5651, 2019.

\bibitem{ni2019elastic}
T.~Ni, L.~Xie, H.~Zheng, E.~K. Fishman, and A.~L. Yuille, ``Elastic boundary
  projection for 3d medical image segmentation,'' in \emph{Proceedings of the
  IEEE Conference on Computer Vision and Pattern Recognition}, 2019, pp.
  2109--2118.

\bibitem{borgefors1986distance}
G.~Borgefors, ``Distance transformations in digital images,'' \emph{Computer
  vision, graphics, and image processing}, vol.~34, no.~3, pp. 344--371, 1986.

\bibitem{clevert2015fast}
D.-A. Clevert, T.~Unterthiner, and S.~Hochreiter, ``Fast and accurate deep
  network learning by exponential linear units (elus),'' \emph{arXiv preprint
  arXiv:1511.07289}, 2015.

\bibitem{he2016identity}
K.~He, X.~Zhang, S.~Ren, and J.~Sun, ``Identity mappings in deep residual
  networks,'' in \emph{European conference on computer vision}.\hskip 1em plus
  0.5em minus 0.4em\relax Springer, 2016, pp. 630--645.

\bibitem{kingma2014adam}
D.~P. Kingma and J.~Ba, ``Adam: A method for stochastic optimization,''
  \emph{arXiv preprint arXiv:1412.6980}, 2014.

\bibitem{badrinarayanan2017segnet}
V.~Badrinarayanan, A.~Kendall, and R.~Cipolla, ``Segnet: A deep convolutional
  encoder-decoder architecture for image segmentation,'' \emph{IEEE
  transactions on pattern analysis and machine intelligence}, vol.~39, no.~12,
  pp. 2481--2495, 2017.

\end{thebibliography}

\end{document}